\documentclass[11pt,a4paper]{article}
\usepackage[hyperref]{acl2019}
\usepackage{times}
\usepackage{latexsym}
\usepackage[mongolian,english]{babel}
\usepackage{xurl}

\usepackage{booktabs}
\usepackage{url}
\usepackage{enumitem}

\aclfinalcopy 

\title{Using Contextual Information for Sentence-level Morpheme Segmentation}

\author{Prabin Bhandari \\
    Department of Computer Science \\
    George Mason University\\
  \texttt{pbhanda2@gmu.edu} \\\And
  Abhishek Paudel \\
    Department of Computer Science \\
    George Mason University\\
  \texttt{apaudel4@gmu.edu} \\}

\date{}

\begin{document}
\maketitle

\begin{abstract}

Recent advancements in morpheme segmentation primarily emphasize word-level segmentation, often neglecting the contextual relevance within the sentence. In this study, we redefine the morpheme segmentation task as a sequence-to-sequence problem, treating the entire sentence as input rather than isolating individual words. Our findings reveal that the multilingual model consistently exhibits superior performance compared to monolingual counterparts. While our model did not surpass the performance of the current state-of-the-art, it demonstrated comparable efficacy with high-resource languages while revealing limitations in low-resource language scenarios.

\end{abstract}

\section{Introduction} \label{sec:introduction}
The problem of morpheme segmentation deals with decomposing a word into a sequence of morphemes that represent the smallest meaningful unit of words like prefixes, suffixes, and root words. For example, the word \emph{pokers} can be decomposed into its morphemes as \emph{poke@@er@@s}, where \emph{@@} represents separations between morphemes. 
Recent advances in morpheme segmentation focus mostly on word-level segmentation \citep{batsuren2022sigmorphon, peters2022beyond} and do not take the context of the word in the sentence into consideration. Instead, we focus on the task of sentence-level morpheme segmentation in which the context of the word in a sentence is taken into account for morpheme segmentation.

The SIGMORPHON 2022 Shared Task on Morpheme Segmentation \citep{batsuren2022sigmorphon} formulates the task of morpheme segmentation into two subtasks: word-level segmentation and sentence-level segmentation. Although many submissions in this task show significant improvements over baselines on both subtasks \citep{rouhe2022morfessor_auuh, wehrli2022cluzh, peters2022beyond}, most submitted approaches for sentence-level segmentation subtask ignore the context of words by design treating the problem as word-level morpheme segmentation, i.e. they treat each word in a sentence as a separate example. 
However, in many languages, the context of a word might provide a piece of meaningful information to disambiguate the morphology of the word \citep{batsuren2022sigmorphon} and help improve morpheme segmentation.
Consider the following sentences in Mongolian and their corresponding morpheme segmentation:
\begin{enumerate}
    \item \foreignlanguage{mongolian}{Гэрт эмээ хоол хийв.} $\rightarrow$ \foreignlanguage{mongolian}{Гэр @@т эмээ хоол хийх @@в.}
    \item \foreignlanguage{mongolian}{Би өдөр эмээ уусан.} $\rightarrow$ \foreignlanguage{mongolian}{Би өдөр эм @@ээ уух @@сан.}
\end{enumerate}

In the above examples, the word \foreignlanguage{mongolian}{эмээ} has different meanings in each sentence. In the first sentence, it means \emph{grandmother} and is not segmentable. However, in the second one, it means \emph{medicine} and is segmentable as \foreignlanguage{mongolian}{эм @@ээ}. Therefore, the context of a word in a sentence is an important factor for determining how the word should be segmented into its morphemes.

We focus on the task of sentence-level morpheme segmentation while taking the whole sentence in context and implement a sequence-to-sequence transformer model \citep{vaswani2017attention} inspired by DeepSPIN-3 \citep{peters2022beyond} which is the winner of word-level morpheme segmentation in the SIGMORPHON 2022 Shared Task on Morpheme Segmentation \citep{batsuren2022sigmorphon}.
While most existing methods treat sentence-level morpheme segmentation as a zero-shot solution of word-level morpheme segmentation \citep{batsuren2022sigmorphon}, we treat each sentence as a whole as one training example to preserve its context and treat the problem as a sequence to sequence generation task. The sentence-level dataset provided for the shared task consists of three languages: Czech, English, and Mongolian.

We perform various experiments with monolingual and multilingual sequence-to-sequence transformer models and show that the multilingual model generally performs better over monolingual models, especially for low-resource languages. Additionally, we experiment with data augmentation in which we increase the training dataset by combining samples for word-level morpheme segmentation. We also experiment with upsampling the sentence-level data given the lack of enough training data for low-resource languages like Czech and Mongolian. Although we are not able to outperform the winners of the shared task, our results are close for high resource languages like English (F1 score: 95.10) but relatively underperform for low-resource languages like Czech (F1 score: 75.79) and Mongolian (F1 score: 72.54).

\section{Approach} \label{sec:approach}
We implement sequence to sequence transformer model similar to DeepSPIN-3 \citep{peters2022beyond} for sentence-level morpheme segmentation while treating each sentence as one training example.
Our implementation is done using fairseq \citep{ott2019fairseq}.
This sequence-to-sequence task setting for sentence-level morpheme segmentation is similar to machine translation settings but differs in the sense that the source language corresponds to sentence examples and the target language corresponds to the sentence in which the words are segmented into their corresponding morphemes.
We train monolingual and multilingual models with the same transformer architecture.
The overall approach for both monolingual and multilingual models are described below.

\subsection{Tokenization}
We use Google's SentencePiece tokenizer with subword regularization using Unigram Language Model (ULM) \citep{kudo2018subword}.
ULM is a top-down technique where the model is initially initialized with a vast vocabulary of overlapping sub-words and a score is generated based on expectation maximization for each sub-word.
Up until the anticipated vocabulary size is attained, the lowest-scoring sub-words are trimmed.

The SentencePiece tokenizer is trained using both source and target data, and all tokens for sentence-level data are pre-computed before training as expected by fairseq's training workflow.
The vocabulary size for English is chosen to be 6000 to 8000 for English and 1000 to 5000 for Czech and Mongolian depending upon the data augmentation which could allow for higher vocabulary sizes.
Additionally, for multilingual model, we use the vocabulary size to be 9000 to 12000 so as to take into account a large number of diverse tokens present in multiple languages.

\subsection{Model Architecture}
Our transformer model \citep{vaswani2017attention} consists of 6 encoder and 6 decoder blocks with 8 multi-head attentions.
We use an embedding size of 256 and dropout of 0.3 determined through hyperparameter tuning on the dev set.
The size of each feed-forward layer is 1024 which is also determined through hyperparameter tuning on the dev set.
We use identical architecture for all three languages: Czech, English, and Mongolian.

\subsection{Training}
We train for a maximum of 400,000 updates and perform early stopping based on the validation loss. 
We use learning rate warm-up for 4000 steps for different languages determined through hyperparameter tuning.
We also use inverse square root learning rate scheduling during our training. We use a batch size of 8192 for all the experiments.

Our model is trained using entmax-loss \citep{peters2019sparse} with alpha of 1.5 similar to \citet{peters2022beyond} since it seemed to show better performance compared to cross-entropy loss in our preliminary experiments.
The entmax loss is a general family of loss functions encompassing the cross-entropy loss as a special case (when alpha=1). Using values greater than 1 for alpha, the entmax loss allows for sparse gradients with non-zero values occurring only on the gold label and/or other labels that receive non-zero probability. This means that completely-peaked probability distributions are possible, in contrast to the denser distributions observed at alpha=1."
This makes entmax loss a better choice over cross-entropy loss in many applications.
However, entmax loss is computationally more expensive to compute than cross-entropy loss.
We therefore use existing implementation of entmax loss rather than implementing it from scratch.

\subsection{Inference and Evaluation}
For generating morpheme segmentation for each sentence, we use beam search with a beam size of 5. 
The tokenizer trained using the training data is used to tokenize sentences during inference. 
Additional postprocessing is done to clean up raw output from the model to obtain final segmentation. 
We use F1 score as our primary metric to evaluate the morpheme segmentation at the sentence-level.
This means that the generated segmentation for a sentence is considered to be correct only if all the words in that sentence is segmented correctly.
Precision, recall and Levenshtein distance, although computed, have not been reported in this paper.

\begin{table*}[t]
    \centering
    \begin{tabular}{lrrr}
        \toprule
                & Czech & English & Mongolian \\
        \midrule
        Monolingual Sequence-to-Sequence Transformer    & 22.62 & 87.94   & 21.00     \\
        \qquad + Upsampling         & 37.58 & 88.62   & 39.40     \\
        \qquad + Word Augmentation  & 25.87 & 92.46   & 16.76     \\
        \qquad + Word Augmentation + Upsampling  & 46.62 & \textbf{95.10}   & 38.00     \\
        \midrule
        Multilingual Sequence-to-Sequence Transformer    & 66.77 & 88.54   & 68.12     \\
        \qquad + Word Augmentation + Upsampling  & \textbf{75.79} & 92.93   &\textbf{ 72.54}     \\
        \midrule
        Official highest score in \citet{batsuren2022sigmorphon}    & 91.99 &  96.31  &  82.88    \\
        DeepSPIN-sent \citep{peters2022beyond}    & \emph{93.23} & \emph{98.24} & \emph{83.59}       \\

        \bottomrule
    \end{tabular}
    \caption{F1 scores obtained for sentence-level test sets for Czech, English and Mongolian languages in monolingual and multilingual settings (top two blocks) along with highest scores of official shared task \citep{batsuren2022sigmorphon} winners (AUUH\_B \citep{rouhe2022morfessor_auuh} for English and CLUZH-3 \citep{wehrli2022cluzh} for Czech and Mongolian ) and DeepSPIN-3 (unofficial highest scorer). The bold numbers are highest scores for our implementation, and italicized numbers are highest overall scores.}
    \label{tab:results}
\end{table*}

\begin{table}
    \centering
    \begin{tabular}{lrrr}
         \toprule
         Language &  train & dev & test\\
         \midrule
         Czech &  1000 & 500 & 500\\
         English &  11007 &  1783 & 1845\\
         Mongolian &  1000 & 500 & 500\\
         \bottomrule
    \end{tabular}
    \caption{Break down of sentence-level morpheme segmentation for Czech, English and Mongolian in train, dev and test set.}
    \label{tab:sentence_dataset}
\end{table}

\section{Experiments}

We evaluate our approach on the sentence-level morpheme segmentation dataset provided by \citet{batsuren2022sigmorphon}.
This dataset consists of train, dev and test samples in Czech, English and Mongolian languages as shown in Table \ref{tab:sentence_dataset}.
We experiment with monolingual models (separate models for each language) and multilingual model (one model for all languages) that are discussed below.

\subsection{Monolingual Experiments}
We experiment with monolingual models in which we train three different models separately on Czech, English and Mongolian datasets.
We conduct few different experiments using the sequence to sequence transformer model as described in \ref{sec:approach} using only the sentence-level dataset, augmenting with word dataset and upsampling the dataset.

\subsubsection{Sentence-level Dataset Only} \label{sec:mono_sentence}
In this experiment, we take the sentence-level dataset in three languages and train the sequence to sequence transformer model for each of these languages.
The results are shown in the Table \ref{tab:results} (second row).
We observe that for high resource language like English, the F1 score is 87.94 while drops significantly for low-resource language likes Czech (F1 score: 22.62) and Mongolian (F1 score: 21.00).

\subsubsection{Data Upsampling} \label{sec:upsampling}
Since there are not many training examples especially in Czech and Mongolian, the sentence-level training set is upsampled 100 times for Czech and Mongolian and 10 times for English.
The results with data upsampling are shown in Table \ref{tab:results} (third row).
We observe that the results for all three languages improve but the gap between English, which is a high-resource language, and low-resource languages like Czech and Mongolian is still large.

\subsubsection{Word-level Dataset Augmentation}

\begin{table}
    \centering
    \begin{tabular}{lr}
         \toprule
         Language &  Number of Samples\\
         \midrule
         Czech & 38682\\
         English & 577374\\
         Mongolian & 18966\\
         \bottomrule
    \end{tabular}
    \caption{Break down of word-level morpheme segmentation dataset for Czech, English and Mongolian used to augment sentence-level dataset.}
    \label{tab:word_dataset}
\end{table}

Another way to increase the number of training data we employ is to augment the sentence-level training set with word-level dataset from their corresponding languages. 
The results with word-level datset augmentation are shown in Table \ref{tab:results} (fourth row).
We observe that augmenting the word-level dataset significantly improves the F1 score for English to 92.46 compared to 87.94 obtained without any upsampling or word-level dataset augmentation.
However, there is only small improvement for Czech language (from 22.62 without any upsampling or word-level dataset augmentation to 25.87 in this case).
Surprisingly, augmenting the word-level dataset reduced the performance for Mongolian language.
This could be because of the prevalence of context-dependent morpheme segmentation examples like the one discussed in Sec. \ref{sec:introduction}.
We believe that understanding the core reason for this drop in performance might require knowledge of Mongolian language's morphology.

\subsubsection{Word Augmentation with Upsampling}
We combine word-level data augmentation and then perform upsampling similar to Sec. \ref{sec:upsampling} and use this dataset to train the transformer model.
The results are shown in Table \ref{tab:results} (fifth row).
We observe that this achieves the highest overall F1 score of 95.10 for English. We also see improvements for Czech with F1 scores of 46.62.
Although there is improvement for Mongolian with F1 score of 38.00 compared to word-level dataset augmentation, this is still lower than F1 score of 39.40 obtained with upsampling, and therefore this improvement is most likely only contributed by upsampling rather than word-level dataset augmentation which most likely slightly hampered the performance instead.

\subsection{Multilingual Experiments}
We experiment with multilingual model in which we concatenate training datasets from Czech, English and Mongolian into one large multilingual dataset and train one tokenizer model and one sequence-to-sequence transformer on this multilingual dataset. 
We do not use any language identifier token to identify the samples from different languages and instead perform simple concatenation only. 
We also experiment with word-level dataset augmentation and then perform upsampling on this augmented multilingual dataset since this approach demonstrated better performance with monolingual models.
The results are shown in Table \ref{tab:results} (middle block).

We observe that the multilingual model outperforms monolingual models for Czech and Mongolian in which the F1 scores are 75.79 and 72.54 respectively. This shows that Czech and Mongolian languages significantly benefited from multilingual settings.
For English, the performance drops slightly to 92.93 in multilingual settings from 95.10 in monolingual settings. Overall, we observe that low-resource languages see much more performance improvements in multilingual settings as expected.

\section{Related Work}

Many NLP applications require tokenization of words into subwords. However, most subword tokenization approaches like Byte-Pair Encoding (BPE) \citep{gage1994new_bpe, sennrich2016neural_bpe} and Unigram Language Model (ULM) \citep{kudo2018subword} ignore the morphophology of words during tokenization and instead tokenize words based on statistical co-occurence of subwords.
\citet{batsuren2022sigmorphon} proposed SIGMORPHON 2022 Shared Task on Morpheme Segmentation citing the necessity of linguistically motivated subword tokenization in which many approaches demonstrated state-of-the-art performance over existing baseline systems like WordPiece \citep{schuster2012japanese_wordpiece}, ULM \citep{kudo2018subword} and Morfessor2 \citep{virpioja2013morfessor}.

For example, DeepSPIN-2 and DeepSPIN-3 \citep{peters2022beyond} collectively demonstrate a superior performance on word-level morpheme segmentation on all 9 languages considered on the shared task.
AUUH \citep{rouhe2022morfessor_auuh}, CLUZH \citep{wehrli2022cluzh} and others also show significantly improved performance over baselines for both subtasks.
However, most submitted approaches ignore the context of words in a sentence for sentence-level morpheme segmentation.

Our approach formulates sentence-level morpheme segmentation by considering sentence as a whole to preserve the context (including the unofficial highest scorer DeepSPIN-sent) and performs morpheme segmentation without separating the sentence into individual words before morpheme segmentation.

\section{Conclusion and Future Work}
Our results show that our approach with multilingual model shows better performance over monolingual models.
Additionally, the performance improvement is more significant for low-resource languages like Czech and Mongolian than English for which comparatively large amount of training data was available.
We also show that word-level dataset augmentation and data upsampling show improved performance particularly for low resource languages. In this regard, we emphasize that simple techniques like data upsampling could serve as a first step for low-resource languages in some NLP tasks.
Overall, although our results do not outperform state of the art, they are competitive for high-resource languages like English and have the potential for improved performance upon tackling issues with regards to the lack of enough data for low-resource languages like Czech and Mongolian.

Future directions of this research could try to tackle lack of enough data for low-resource languages by investigating techniques like semi-supervised learning in which a trained model is used to generate labels for newly collected data and then integrate this newly labelled data to expand the training data if the predicted probabilities for the labels cross certain threshold. This could help generate new labelled data that could then be used to train new models. Additionally, our multilingual model do not distinguish between language samples. Adding an additional language identifier token in data based on language could potentially help improve the performance of multilingual model further.

\section*{Acknowledgements}
We would like to thank Antonios Anastasopoulos and Ziyu Yao for their guidance and mentorship. We are also grateful to Ben Peters for making their code open source for us to build upon, and for their thoughtful feedback on an earlier version of this paper.

\bibliographystyle{acl_natbib} 
\bibliography{refs} 

\end{document}